\pdfoutput=1
\documentclass[11pt]{article}
\usepackage[]{ACL2023}

\usepackage{times}
\usepackage{latexsym}
\usepackage{multirow}
\usepackage{graphicx}
\usepackage{framed}
\usepackage{arydshln}
\usepackage[normalem]{ulem}
\usepackage{amssymb}

\usepackage[T1]{fontenc}

\usepackage[utf8]{inputenc}

\usepackage{microtype}

\usepackage{booktabs}

\usepackage{inconsolata}
\usepackage{makecell}
\usepackage{amsmath}
\usepackage{array}
\newcolumntype{C}[1]{>{\raggedright\arraybackslash}p{#1}}

\DeclareMathOperator*{\argmax}{argmax}

\definecolor{OliveGreen}{rgb}{0,0.6,0}

\title{gTBLS: Generating Tables from Text by Conditional Question Answering}

\author{Anirudh Sundar, Christopher Richardson, Larry Heck \\
  AI Virtual Assistant (AVA) Lab \\
  Georgia Institute of Technology \\
  \texttt{\{asundar34,crichardson8,larryheck@gatech.edu\}} \\}

\begin{document}
\maketitle
\begin{abstract}
Distilling large, unstructured text into a structured, condensed form such as tables is an open research problem. One of the primary challenges in automatically generating tables is ensuring their syntactic validity. Prior approaches address this challenge by including additional parameters in the Transformer's attention mechanism to attend to specific rows and column headers. In contrast to this single-stage method, this paper presents a two-stage approach called Generative Tables (\mbox{gTBLS}). The first stage infers table structure (row and column headers) from the text. The second stage formulates questions using these headers and fine-tunes a causal language model to answer them. Furthermore, the \mbox{gTBLS} approach is amenable to the utilization of pre-trained Large Language Models in a zero-shot configuration, presenting a solution for table generation in situations where fine-tuning is not feasible. \mbox{gTBLS} improves prior approaches by up to 10\% in BERTScore on the table construction task and up to 20\% on the table content generation task of the E2E, \mbox{WikiTableText}, WikiBio, and RotoWire datasets.
\end{abstract}

\section{Introduction}

An important challenge in Natural Language Processing is summarization, distilling large, unstructured texts into a condensed form while preserving factual consistency. There has been substantial work summarizing news articles, medical information, and conversational dialogue \cite{nallapati2016abstractive, see-etal-2017-get, shang-etal-2018-unsupervised,joshi-etal-2020-dr, chen-yang-2020-multi}. However, these efforts focus on transforming unstructured text into shorter yet unstructured forms. Compiling unstructured knowledge sources into structured forms such as tables remains an open research problem.

Organizing information into tables provides several advantages compared to unstructured paragraphs \cite{tang_struc-bench_2023}. Tabular information is more efficient, utilizing row and column headers to reduce redundancy. 
Additionally, the structured presentation simplifies the task of comparing different sources of information, especially when dealing with quantitative data. However, manually creating tables from text is time-consuming and necessitates an automated approach. 

\begin{figure}[t]
    \centering
    \includegraphics[width=\columnwidth]{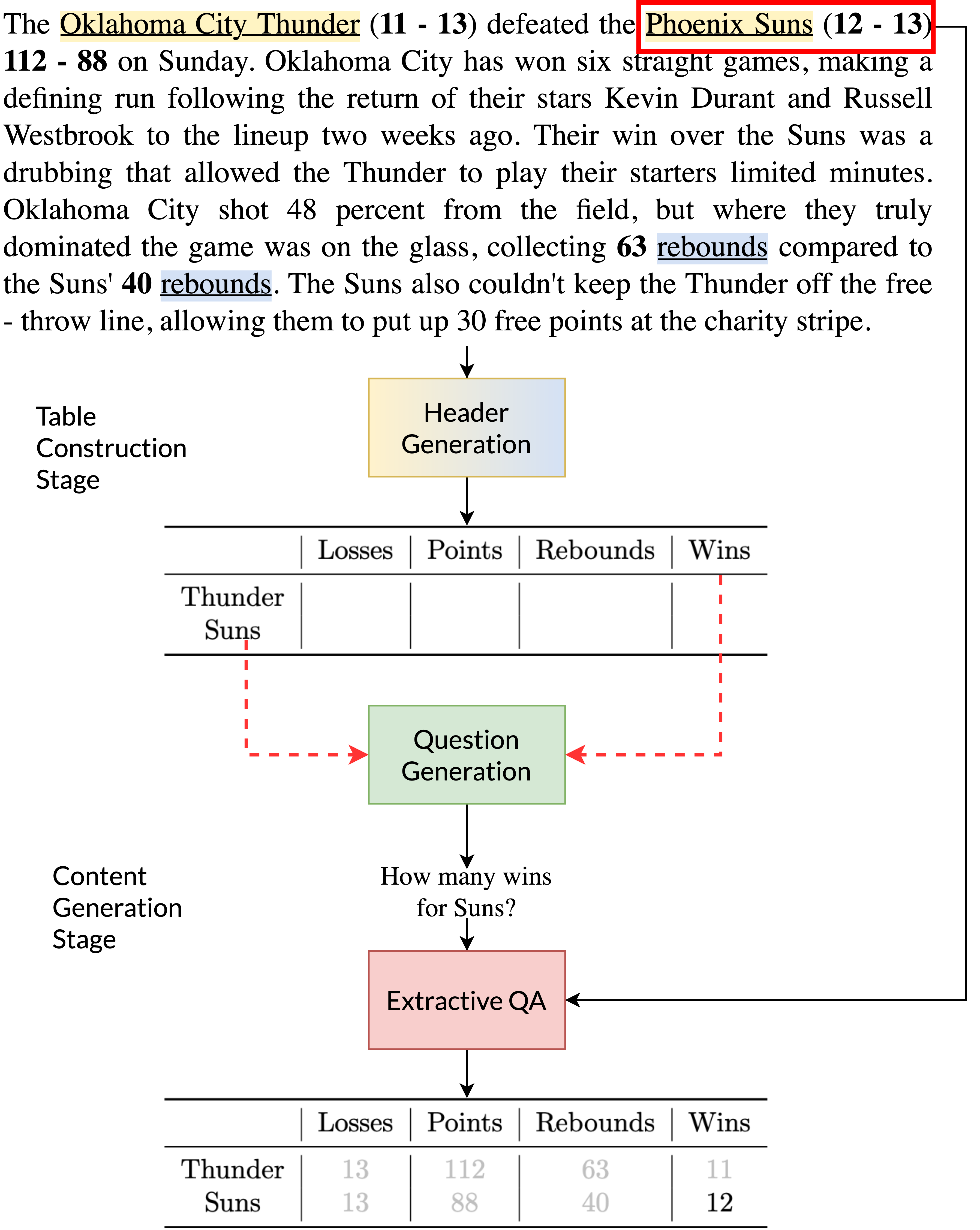}
    \caption{Overview of Generative Tables (\mbox{gTBLS}). \mbox{gTBLS} uses a two stage approach to condense textual information into structured tables.}
    \vspace{-5pt}
    \label{fig:enter-label}
\end{figure}

Driven by the success of Large Language Models (LLMs) on sequence-to-sequence natural language tasks, recent work explored the automatic generation of structured knowledge from unstructured text \cite{wu-etal-2022-text-table, pietruszka_stable_2022}. A primary challenge in this automatic table generation task lies in ensuring their syntactic validity. Every row and column in a table must contain the same number of cells, with row and column headers delineating relationships between cells. Failing to adhere to this constraint invalidates the structure of the table and the information presented. Prior work addresses this constraint by including additional parameters like row and column relation embeddings \cite{wu-etal-2022-text-table} or positional bias \cite{pietruszka_stable_2022} to get the model to attend to header cells while generating content.

In contrast, we propose Generative Tables (\mbox{gTBLS}) \footnote{Our code will be released with the camera-ready version} as shown in Figure \ref{fig:enter-label}, a novel, two-stage approach to condense unstructured textual information into structured tables. 
While prior work \cite{pietruszka_stable_2022, wu-etal-2022-text-table} relied on learning the implicit relationship between header cells and content cells using additional parameters in the Attention mechanism \cite{vaswani2017attention}, \mbox{gTBLS} makes this process explicit by splitting the task of table generation from text into two stages - Table Construction and Table Content Generation. 
Table Construction infers table structure (row and column headers) from text. Table Content Generation uses the generated headers to formulate questions. An LLM is then fine-tuned to answer these questions using the textual paragraph as evidence. Alternatively, to underscore the advantages of the modular two-stage approach in \mbox{gTBLS}, one can utilize \textit{off-the-shelf} LLMs in a zero-shot configuration to perform the question-answering in the second stage.  

The advantages of the modular two-stage approach are manifold:
\begin{enumerate}
    \item By splitting the task of automatic table generation into Table Construction and Content Generation, \mbox{gTBLS} ensures all tables are syntactically valid (equal number of cells across rows and columns), resulting in up to 57\% reduction in error rates compared to a sequence-to-sequence approach with no constraints.
    \item By making the relationship between header cells and content explicit through question answering, \mbox{gTBLS} achieves up to 20\% improvement in BERTScores on the table generation task proposed by \cite{wu-etal-2022-text-table}.
    \item By demonstrating the ability of instruction fine-tuned LLMs to perform Table Content Generation, the question-answering reformulation can utilize larger LLMs to achieve parity with fine-tuning approaches, presenting a solution for table generation in situations where fine-tuning is not feasible.
    \item By reformulating the table generation task as question answering, new evidence can be incorporated into existing tables using \mbox{gTBLS} without regeneration of the entire table.
\end{enumerate}
This paper is structured as follows: 
Section \ref{sec:related_work} reviews related work that addresses the challenge of generating structured content from textual paragraphs. Section \ref{sec:tab_gen_as_QA} describes our novel approach \mbox{gTBLS}, which reformulates table generation as conditional question answering. Section \ref{sec:experiments} describes the dataset, experimental procedure, and results. We conclude the paper in Section~\ref{sec:conclusions} and outline limitations in Section~\ref{sec:limitations}.

\section{Related Work}
\label{sec:related_work}

Early research on tabular generation focused on discriminative techniques. \citet{branavan-etal-2007-generating} used a tree-based method to infer a table of contents from documents, while \citet{aramaki_text2table_2009} treated tabular generation as a multi-label classification problem, with predefined headers.

More recent neural approaches for table generation have utilized Generative Adversarial Networks (GANs) to synthesize tabular data from existing datasets \cite{xu_synthesizing_2018, park_data_2018, chen_faketables_2019}. Similarly, research in generating structured information, like knowledge graphs and entities from text, has also been explored \cite{hakkani-tur_weakly-supervised_2013, luan_multi-task_2018, deng-etal-2021-structure, lu-etal-2022-unified}.

Recent work directly addressing text-to-table generation includes \cite{wu-etal-2022-text-table}, \cite{pietruszka_stable_2022}, and \cite{tang_struc-bench_2023}. \citet{wu-etal-2022-text-table} proposed modifying the Transformer decoder's attention mechanism, incorporating row and column relation embeddings to capture header and non-header cell relationships. \citet{pietruszka_stable_2022} utilize learnable bias parameters to encode relative cell positions. Finally, \citet{tang_struc-bench_2023} employed structure-aware instruction-tuning to fine-tune LLMs to generate tables.

However, the end-to-end neural approaches are limited by the fact that the entire architecture needs to be re-trained to leverage newer and better LLMs and relies on learning inter-cell relationships implicitly in a neural space. 
To overcome this limitation, we leverage prior work motivating the advantages of reformulating various NLP tasks as Question Answering. \citet{levy-etal-2017-zero} addresses relation extraction by associating natural language questions with each relation slot. \citet{mforms_arxiv} and \citet{mforms2024} present an approach to form-filling
by reformulating the task as multimodal natural language Question Answering. \citet{li-etal-2020-unified} address Named Entity Recognition as Machine Reading Comprehension. \citet{9413810} and \citet{du-etal-2021-qa} map Natural Language Understanding tasks to few-shot and zero-shot Question Answering. \citet{fuisz2022improved} provide evidence to the advantages of Question Answering for the Slot Labeling task. \citet{nakamura-etal-2022-hybridialogue} and \citet{sundar-heck-2023-ctbls} demonstrate approaches to answer questions grounded in tabular content.

Motivated by these methods, our approach, \mbox{gTBLS}, uses a two-stage process splitting the task into table structure construction and table content generation to capture inter-cell relationships and adhere to tabular constraints.

\section{Table Generation as Question Answering}
\label{sec:tab_gen_as_QA}

The foundation of Generative Tables (\mbox{gTBLS}) is a two-stage approach to table generation that disentangles structure generation and information retrieval. While LLMs have demonstrated success on text generation and information retrieval independently, utilizing them to generate structured knowledge is more complex. Rows and columns impose structure requirements during inference.  LLM-based methods that generate tables sequentially (e.g., row-by-row or column-by-column) face a critical challenge: the number of cells generated in the initial row or column determines the structure of the entire table. Failing to adhere to these constraints results in structurally invalid tables. \mbox{gTBLS} addresses this issue by first employing a Table Construction stage to identify row and column headers from natural language text to construct an empty table with headers (represented by the upper portion of Figure \ref{fig:enter-label}). Then, the Table Content Generation stage uses the identified headers to fill cell contents with synthetically generated QA pairs, ensuring the validity of all generated tables  (represented by the bottom half of Figure \ref{fig:enter-label}).

\subsection{Table Construction}
\label{subsec:table_construction}
The Table Construction stage is formulated as Conditional Text Generation where the task is to generate a sequence of headers $\{h_{1} \ldots h_{n}\}$ from the input textual paragraph $t$. In this stage, \mbox{gTBLS} utilizes an encoder-decoder language model to generate row and column headers in a supervised approach. During training, the model is trained to extract row and column headers using teacher-forcing. Since encoder-decoder models produce text sequentially, the target is a sequence of concatenated headers separated by a <SEP> token (for example, Rebounds <SEP> Assists <SEP> Points). The language model is fine-tuned to generate the concatenated header sequence autoregressively, conditioned on the textual paragraph, by maximizing the causal language modeling objective 
\begin{equation}
    \argmax_h p(h_i|\{h_1 \ldots h_{i-1}\},t).
    \label{eq:eq_1}
\end{equation}
During inference, the model generates a sequence of headers based solely on the textual input.

\subsection{Table Content Generation}
\label{subsec:table_content_generation}
The Table Content Generation stage generates synthetic QA pairs over the skeleton of the table constructed in the previous stage. Using the generated rows and columns from Table Construction, \mbox{gTBLS} formulates a question, the answer to which is the cell content. A separate question is formulated for each combination of row and column header in the format `What is the \{Column value\} for \{Row value\}?'. For example, given the row header `Suns' and the column header `Wins', the formulated question is `What is the number of Wins for Suns?', as shown in the bottom half of Figure \ref{fig:enter-label}. An encoder-decoder LLM is either deployed in a zero-shot configuration or fine-tuned to answer this question using the textual input as evidence. Given row and column headers $h_{row}$ and $h_{col}$, the objective during fine-tuning is to maximize the probability of the correct response $r$ to the question $q$ given the paragraph $t$
\begin{equation}
    \argmax_r p(r|q(h_{row}, h_{col}),t).
    \label{eq:eq_2}
\end{equation}
The zero-shot approach utilizes instruction fine-tuned encoder-decoder LLMs to generate answers to the formulated questions. At inference, in contrast to prior work that generates tables row-by-row or column-by-column, batching the questions corresponding to a single table allows all cell content to be generated simultaneously. 

\section{Experimental Results}
\label{sec:experiments}

\subsection{Text-to-Table Datasets}

\begin{table}[t]
    \centering
    \begin{tabular}{c|c|c|c}
    \toprule 
        Dataset & Train & Valid & Test  \\
        \midrule
        E2E & 42.1k & 4.7k & 4.7k \\
        WikiTableText & 10k & 1.3k & 2.0k \\
        WikiBio & 582.7k & 72.8k & 72.7k \\
        RotoWire & 3.4k & 727 & 728 \\
        \bottomrule
    \end{tabular}
    \caption{Statistics of the E2E, WikiTableText, WikiBio, and RotoWire datasets, number of samples across splits}
    \label{tab:ds_statistics}
    \vspace{-5pt}
\end{table}

\citet{wu-etal-2022-text-table} propose four datasets for the text-to-table task by inverting datasets created for the dual problem of generating textual descriptions from tables. Each dataset consists of textual paragraphs paired with tabular information summarizing content in the text. Dataset statistics are described in Table \ref{tab:ds_statistics}. Each dataset is described below.

E2E \cite{novikova_e2e_2017} concerns restaurant descriptions, requiring summarization of information into tables with descriptors like restaurant name, customer rating, and location.

WikiTableText (WTT) \cite{bao_table--text_2018}, sourced from Wikipedia, consists of natural language descriptions generated from tabular data across various topics. 

WikiBio \cite{lebret-etal-2016-neural} comprises introductions of individuals from Wikipedia alongside tabular summaries extracted from the same page's information box.
In contrast to E2E, the table headers in the WikiTableText and WikiBio datasets vary widely across data samples. 

RotoWire (RW) \cite{wiseman-etal-2017-challenges} contains NBA game reports and two separate tables summarizing team and player statistics. 

While E2E, WikiTableText, and \mbox{WikiBio} consist of single-column tables, RotoWire is a more challenging dataset with multi-row, multi-column tables, necessitating strict adherence to equal cell counts across rows and columns. The RotoWire dataset consists of two splits - Team and Player statistics. The row headers represent teams and players mentioned in the textual paragraph while the column headers contain information regarding various statistics such as assists, rebounds, and points. The specific headers for each data sample vary based on the information provided in the textual description. Furthermore, E2E, WikiTableText, and WikiBio consist of tables with textual content while the RotoWire datasets contain numerical data. Example textual paragraphs and associated tables from each dataset are presented in Appendix \ref{sec:dataset_examples}. 

\begin{table}[t]
\addtolength{\tabcolsep}{-1pt}
    \centering
    \begin{tabular}{c|c|c|c}
    \toprule
         \multirow{2}*{Dataset} & \multirow{2}*{Model} & \multicolumn{2}{c}{Header Cell}  \\ 
         & & F1 & BERTScore\\
         \midrule
          \multirow{2}*{E2E} & Wu et al. & \textbf{99.63}  & 99.88 \\ 
          & \mbox{gTBLS} & \textbf{99.61} & \textbf{99.98}  \\ 
         \midrule 
         \multirow{2}*{WikiTableText} & Wu et al. & \textbf{78.16}  & 95.68 \\
          & \mbox{gTBLS} & 74.75  & \textbf{99.37}  \\
         \midrule 
         \multirow{2}*{Wikibio} & Wu et al. & \textbf{80.52}  & 92.60  \\
          & \mbox{gTBLS} & \textbf{80.53} & \textbf{98.72}  \\
         \bottomrule
    \end{tabular}
        
        \vspace{5pt}
        
        \addtolength{\tabcolsep}{-1pt}

        \begin{tabular}{c|c|c| c|c|c}
    \toprule 
         \multirow{2}*{Dataset} & \multirow{2}*{Model} & \multicolumn{2}{c|}{Row Header} & \multicolumn{2}{c}{Col Header}   \\ 
         & & F1  & BS & F1 & BS \\
         \midrule 
         \multirow{3}*{\makecell{RW \\ Team}} & Wu et al. & 94.97  & 97.51 & 86.02  & 89.05  \\
        & STable & 94.97 & 97.80 & \textbf{88.90}  & 88.70  \\ 
         & \mbox{gTBLS} & \textbf{96.21}  & \textbf{99.93} & 85.47  & \textbf{98.54} \\
         
         \midrule 
         \multirow{3}*{\makecell{RW \\ Player}} & Wu et al. & 92.31  & 93.71 & 87.78  & 94.41  \\

        & STable & \textbf{93.50}  & 95.10 & \textbf{88.10} & 94.50 \\ 
         
         & \mbox{gTBLS} & 92.66  & {\bf 99.09} & 85.28  & \textbf{99.33} \\
         \bottomrule
    \end{tabular}
    
    \caption{Comparison between the performance of Generative Tables (\mbox{gTBLS}) and the prior state of the art introduced by \citet{wu-etal-2022-text-table} for Table Construction. BS = BERTScore}
    \label{tab:comapre_sota_tab_con}
    \vspace{-10pt}
\end{table}

\begin{table*}[t]
    \centering
    \begin{tabular}{p{0.3\textwidth}|p{0.3\textwidth}|p{0.3\textwidth}}
       \toprule 
       Text & Predicted Headers & Ground Truth \\
       \midrule 
       Michelle Schimel was New York State assemblywoman in Portuguese Heritage Society.  & title, subtitle, name, \textcolor{red}{position} & title, subtitle, name, \textcolor{OliveGreen}{office} \\
       \midrule 
       Sonia Gandhi was awarded as Order of King Leopold by the Government of Belgium in 2006.  & title, subtitle, year, name, \textcolor{red}{awarding body} & title, subtitle, year, name, \textcolor{OliveGreen}{awarding organization} \\
       \midrule 
       The personal best of Maryam Yusuf Jamal in 800 m was 1:59.69. & title, subtitle, \textcolor{red}{event}, \textcolor{red}{time (min)} & title, subtitle, \textcolor{OliveGreen}{distance}, \textcolor{OliveGreen}{mark} \\
    \bottomrule 
    \end{tabular}
    \caption{Differences between the headers predicted by \mbox{gTBLS} and the ground truth headers from WikiTableText}
    \label{tab:headers_pred_diff}
\end{table*}
\subsection{Table Construction}
\textbf{Training}: We fine-tune \texttt{Flan-T5-base} \cite{chung_scaling_2022} to generate headers for the different datasets as per the approach outlined in Section \ref{subsec:table_construction}. The input to the encoder is the textual paragraph and the targets are the sequence of concatenated headers. We fine-tune for 10 epochs with AdamW \cite{loshchilov_decoupled_2019} on 8 Nvidia A40 GPUs and use greedy sampling in the decoding process. Experimenting with multiple runs of non-greedy decoding followed by averaging predictions did not yield noticeably different results. Further hyperparameters are listed in Appendix \ref{sec:hyperparams}. 

\textbf{Evaluation}: To evaluate the generated headers, we compute F1 scores and report the results in Table \ref{tab:comapre_sota_tab_con}.  The F1 score is the harmonic mean of precision and recall of the predicted header cells compared to the ground truth. The F1 scores of our approach achieve parity with or surpass the prior State of the art (SoTA) \cite{wu-etal-2022-text-table}, \cite{pietruszka_stable_2022} on the E2E, WikiBio, and RotoWire Team datasets and is within 4\% relative to the F1 score on WikiTableText and the RotoWire Team and Player datasets.

\begin{table*}
    \centering
    \begin{tabular}{p{0.24\textwidth}|p{0.24\textwidth}|p{0.24\textwidth}|p{0.24\textwidth}}
    \toprule 
    \toprule 
    \multicolumn{4}{{C{0.96\textwidth}}}{\textbf{1. Text}: Leonard Shenoff Randle (born February 12, 1949) is a former Major League Baseball player. He was the first-round pick of the Washington Senators in the secondary phase of the June 1970 Major League Baseball draft, tenth overall.} \\ 
    \multicolumn{4}{{C{0.96\textwidth}}}{\textbf{Generated Table}:} \\ 
    \midrule
    Header & Prediction - ZS & Prediction - FT & Ground Truth \\
    \midrule 
    Name & Leonard Randle & Len Randle & Lenny Randle \\
    Birth Date & \textcolor{red}{February 12, 1949} & \textcolor{OliveGreen}{12 February 1949} & 12 February 1949 \\
    Debut Team & Washington Senators & Washington Senators & Washington Senators \\
    \midrule 
    \midrule 
    \multicolumn{4}{{C{0.96\textwidth}}}{\textbf{2. Text}: John "Jack" Reynolds (21 February 1869 — 12 March 1917) was a footballer who played for, among others, West Bromwich Albion, Aston villa and Celtic. as an international he played five times for Ireland before it emerged that he was actually English and he subsequently played eight times for England. he is the only player, barring own goals, to score for and against England and is the only player to play for both Ireland and England. He won the FA cup with West Bromwich Albion in 1892 and was a prominent member of the successful Aston villa team of the 1890s, winning three English league titles and two FA cups, including a double in 1897.} \\ 
    \midrule
    Header & Prediction - ZS & Prediction - FT & Ground Truth \\
    \midrule 
    Name &  \textcolor{red}{John ``Jack" Reynolds} & \textcolor{OliveGreen}{Jack Reynolds} & Jack Reynolds \\
    Birth Date & 21 February 1869 & 21 February 1869 & 21 February 1869 \\
    Death Date & 12 March 1917 & 12 March 1917 & 12 March 1917 \\
    Full Name & \textcolor{red}{John ``Jack" Reynolds} & \textcolor{OliveGreen}{John Reynolds} & John Reynolds \\
    \midrule 
    \midrule 
    \multicolumn{4}{{C{0.96\textwidth}}}{\textbf{3. Text}: Mississippi State Bulldogs Baseball won Virginia in 2013 at Charlottesville, VA.} \\ 
    \midrule 
    Header & Prediction - ZS & Prediction - FT & Ground Truth \\
    \midrule 
    Title & Bulldogs win Virginia & Bulldogs Win Virginia & Mississippi State Bulldogs Baseball \\
    Subtitle & Bulldogs Baseball wins Virginia & Bulldogs Baseball wins Virginia & Bulldogs in the NCAA tournament \\
    Year & 2013 & 2013 & 2013 \\
    Opponent & Virginia & Virginia & Virginia \\
    Site & \textcolor{red}{Charlottesville, VA} & \textcolor{OliveGreen}{University of Virginia} & University of Virginia \\
    \midrule 
    \midrule 
    \end{tabular}
    
    \caption{Difference between the tables generated by the Zero Shot (ZS) and Fine-Tuned (FT) approaches with respect to the Ground Truth on the WikiBio and WikiTableText datasets}
    \label{tab:difference_answers}
\end{table*}
To understand the difference between the performance of \mbox{gTBLS} and prior SoTA, we analyzed the generated headers. We observed that a number of the header cells in the tables were subjective, with many possible interpretations that were semantically valid. Table \ref{tab:headers_pred_diff} contains sample cases from the WikiTableText dataset highlighting the variety in the possible headers generated for different data samples. From the results, it is evident that though not identical, several headers are semantically equivalent. For example, in the context of politics, the terms `position' and `office' can be used interchangeably. Similarly, `awarding body' and `awarding organization' also convey the same meaning. Finally, `event' and `distance' can both be used to demarcate competitions in athletics.  

Therefore, to underscore the performance of \mbox{gTBLS}, we compute the BERTScore of the generated headers with respect to the reference headers and report results in the second column of Table \ref{tab:comapre_sota_tab_con}. BERTScore \cite{zhang_bertscore_2020} measures token similarity between candidate and reference sentences through contextual embeddings, and captures semantic similarity. Observing the BERTScore results in Table \ref{tab:comapre_sota_tab_con}, \mbox{gTBLS} emerges as the best method across all datasets, achieving a relative improvement of up to 10.6\% with respect to prior work and represents the new SoTA. 

\subsection{Table Content Generation}
\textbf{Training}: The next stage in the \mbox{gTBLS} pipeline is Table Content Generation. This stage generates the cell content following the QA reformulation described in Section \ref{subsec:table_content_generation}. We experiment with both fine-tuning and zero-shot approaches for Table Content Generation. For question-answer fine-tuning experiments, we utilize \texttt{Flan-T5-base} \cite{chung_scaling_2022}. Using teacher-forcing for each cell, we synthesize questions from the corresponding row and column headers. The encoder is provided with the input text paragraph and the question corresponding to a single cell. The decoder then generates the answer to this question. We fine-tune for 10 epochs with AdamW \cite{loshchilov_decoupled_2019} and utilize greedy sampling for decoding. Additional hyperparameters are described in Appendix \ref{sec:hyperparams}. 

To further demonstrate the advantages of the modular two-stage approach, we conduct experiments in a \textit{zero-shot} configuration. Here, we utilize larger encoder-decoder models from the Flan-T5 family, namely, \texttt{Flan-T5-large, Flan-T5-xl}, and \texttt{Flan-T5-xxl} that are already instruction fine-tuned for a number of tasks including extractive question-answering. For the RotoWire datasets, since the cell content is purely numerical, each generated response is processed to extract the first occurrence of a number (e.g. ` Ricky Rubio talled just five points' $\rightarrow$ 5). 

Table \ref{tab:zs_ft_tab_con_gen} reports the performance of different approaches for Table Content Generation on all dataset splits in terms of F1 and BERTScore. The zero-shot approach struggles on the WikiTableText dataset due to the open-ended nature of the questions (questions of the form `What is the title of the table?' have multiple valid responses), represented by the relatively lower exact match scores. In contrast, the zero-shot approach excels on the RotoWire datasets with numerical responses, performing within 6\% relative to the full-fine tuning approach in terms of exact match and nearly identical in terms of BERTScore. Additionally, the two-dimensional structure of the tables in RotoWire helps the zero-shot approach since there is additional context to answer the questions.

\begin{table}[t]
    \centering
    \addtolength{\tabcolsep}{-1pt}
    \begin{tabular}{c|c|c|c|c}
    
    \toprule
        \multirow{2}*{Dataset} & Approach & ZS \includegraphics[height=11pt]{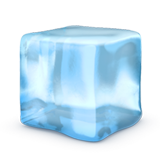} / & \multirow{2}*{F1} & BERT- \\
        & \texttt{Flan-T5-} & FT \includegraphics[height=11pt]{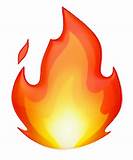} &  & Score\\
        \midrule 
        \multirow{4}*{E2E} & \texttt{large} & \includegraphics[height=11pt]{ice.png} & 83.71 & 94.39 \\
                           & \texttt{xl} & \includegraphics[height=11pt]{ice.png} & \underline{86.11} & \underline{95.33} \\
                           & \texttt{xxl} & \includegraphics[height=11pt]{ice.png}  &  76.72 & 89.28 \\
                           & \texttt{base} & \includegraphics[height=11pt]{fire.jpg} & \textbf{98.29} & \textbf{99.87} \\
        \midrule 
        \multirow{4}*{\makecell{WTT}} & \texttt{large} & \includegraphics[height=11pt]{ice.png} & 37.80 & 87.97 \\
                           & \texttt{xl} & \includegraphics[height=11pt]{ice.png} & 37.90 & 87.71 \\
                           & \texttt{xxl} & \includegraphics[height=11pt]{ice.png} & \underline{38.98} & \underline{88.32} \\
                           & \texttt{base} & \includegraphics[height=11pt]{fire.jpg} & \textbf{72.41} & \textbf{97.96} \\
        \midrule 
        \multirow{4}*{WikiBio} & \texttt{large} & \includegraphics[height=11pt]{ice.png} & 50.79 & 91.23 \\
                           & \texttt{xl} & \includegraphics[height=11pt]{ice.png} & 57.58 & 93.43 \\
                           & \texttt{xxl} & \includegraphics[height=11pt]{ice.png} & \underline{58.69} & \underline{94.26} \\
                           & \texttt{base} & \includegraphics[height=11pt]{fire.jpg} & \textbf{67.45} & \textbf{97.79} \\
        \midrule 
        \multirow{4}*{\makecell{RotoWire \\ Team}} & \texttt{large} & \includegraphics[height=11pt]{ice.png} & 49.41 & 89.69 \\
                           & \texttt{xl} & \includegraphics[height=11pt]{ice.png} & 88.98 & 98.77 \\
                           & \texttt{xxl} & \includegraphics[height=11pt]{ice.png} & \underline{90.28} & \underline{99.88} \\
                           & \texttt{base} & \includegraphics[height=11pt]{fire.jpg} & \textbf{95.94} & \textbf{99.99} \\
        \midrule 
        \multirow{4}*{\makecell{RotoWire \\ Player}} & \texttt{large} & \includegraphics[height=11pt]{ice.png} & 56.28 & 99.39 \\
                           & \texttt{xl} & \includegraphics[height=11pt]{ice.png} & 75.58 & 99.18 \\
                           & \texttt{xxl} & \includegraphics[height=11pt]{ice.png} & \underline{85.48} & \underline{99.77} \\
                           & \texttt{base} & \includegraphics[height=11pt]{fire.jpg} & \textbf{88.75} & \textbf{99.99} \\
        \midrule 
    \end{tabular}
    \caption{Evaluation of the Content Generation Stage in \mbox{gTBLS} - Comparison between Zero Shot (ZS) and Fine-Tuned (FT) approaches.}
    \label{tab:zs_ft_tab_con_gen}
    \vspace{-15pt}
\end{table}

In general, the larger models perform better in a zero-shot configuration but fall short of full fine-tuning of a smaller model. Table \ref{tab:difference_answers} highlights some of the differences between the responses generated by the fine-tuned versus the zero-shot approach. The fine-tuned model is better able to adapt to the format of the references in the ground truth (12 February 1949 vs February 12 1949) and nicknames (Len vs Leonard, Jack vs John ``Jack"). Furthermore, the zero-shot model relies on implicit knowledge obtained during pre-training to make certain inferences during question-answering. While the ground-truth answer refers to the University of Virginia, the zero-shot approach generates the site as Charlottesville, the city where the University is located. Therefore, while the exact match scores from the zero-shot approach are non-competitive with the full-fine tuning, the BERTScore almost achieves parity on WikiBio and the RotoWire datasets. Utilizing larger models from the GPT family \cite{brown2020language} or LLaMA \cite{touvron2023llama} could result in better performance. However, the risk of data snooping is high since the WikiTableText and WikiBio datasets are collected from Wikipedia.

Table \ref{tab:comapre_sota_tab_gen} presents results on the combined task, utilizing the generated headers from the Table Construction stage and the best approach from the Table Content Generation stage. Observing the results, \mbox{gTBLS} emerges as the best method across all datasets, demonstrating up to 20\% relative improvement in BERTScore. In terms of F1, \mbox{gTBLS} achieves parity with prior SoTA on E2E and the RotoWire Player dataset, and demonstrates up to 8.5\% relative improvement on the WikiTableText and RotoWire datasets.

\begin{table}[t!]
\addtolength{\tabcolsep}{-1pt}
    \centering
    \begin{tabular}{c|c|c|c}
    \toprule
         \multirow{2}*{Dataset} & \multirow{2}*{Model} & \multicolumn{2}{c}{Non-Header Cell}  \\ 
         & & F1 & BERTScore\\
         \midrule
          \multirow{2}*{E2E} & Wu et al. & \textbf{97.94}  & 98.57 \\ 
          & \mbox{gTBLS} & \textbf{97.91} & \textbf{99.85}  \\ 
         \midrule 
         \multirow{2}*{WikiTableText} & Wu et al. & 62.71  & 80.74 \\
          & \mbox{gTBLS} & \textbf{68.09}  & \textbf{97.45}  \\
         \midrule 
         \multirow{2}*{Wikibio} & Wu et al. & \textbf{69.71}  & 76.56  \\
          & \mbox{gTBLS} & 67.10 & \textbf{92.53}  \\
         \midrule 
         \multirow{3}*{\makecell{RotoWire \\ Team}} & Wu et al. & 86.31 & 90.80\\
          & \mbox{STable} & 84.70 & 90.30\\
          & \mbox{gTBLS} & \textbf{89.09} & \textbf{97.11}\\
          \midrule 
          \multirow{3}*{\makecell{RotoWire \\ Player}} & Wu et al. & \textbf{86.83} & 88.97\\
          & \mbox{STable} & 84.50 & 90.40\\
          & \mbox{gTBLS} & 86.09 & \textbf{95.61}\\
          
         \bottomrule
    \end{tabular}
    \caption{Comparison between the performance of Generative Tables (\mbox{gTBLS}) and the prior SoTA introduced by \citet{wu-etal-2022-text-table} and \citet{pietruszka_stable_2022} for combined header and content table generation.}
    \label{tab:comapre_sota_tab_gen}
    \vspace{-10pt}
\end{table}

\begin{table*}[t]
    \centering
    \begin{tabular}{c|c|c|c|c|c|c}
        \toprule 
        \multirow{2}*{Dataset} & \multicolumn{2}{c|}{Header F1} & \multicolumn{2}{c|}{Cell F1} &  \multicolumn{2}{c}{Error Rate}  \\
            &  Seq2Seq & \mbox{gTBLS} & Seq2Seq & \mbox{gTBLS} & Seq2Seq & \mbox{gTBLS} \\
            \midrule 
        E2E & \textbf{99.60} & \textbf{99.61} & \textbf{97.94} & \textbf{97.91} & \textbf{0.0\%}& \textbf{0.0\%}\\
        WikiTableText & 69.71 & \textbf{74.75} & 66.61 & \textbf{68.09} & 0.6\% & \textbf{0.0\%}\\
        Wikibio & 76.36 & \textbf{80.53} & 63.51 & \textbf{66.98} & 1.64\% & \textbf{0.0\%}\\
        RotoWire Team & 57.84 & \textbf{90.84} & 51.18 & \textbf{89.09} & 30.9\% & \textbf{0.0\%} \\
        RotoWire Player & 26.34 & \textbf{88.97} & 12.80 & \textbf{86.09} & 57.28\% & \textbf{0.0\%} \\
        \bottomrule 
    \end{tabular}
    \caption{Comparison of F1 scores between sequence to sequence baseline and \mbox{gTBLS}}
    \label{tab:seq_to_seq}
    \vspace{-5pt}
\end{table*}

\subsection{Syntactic Validity}
In Table \ref{tab:seq_to_seq}, we compare \mbox{gTBLS} with a sequence-to-sequence approach that models table generation as conditional generation of a flattened table representation. The sequence-to-sequence baseline uses a single \texttt{Flan-T5-base} model fine-tuned to generate the entire table conditioned on the input text in a single stage. To parse the output as a valid table, the `|' token is used to separate columns and a <NEWLINE> tag separates rows. The sequence-to-sequence baseline is fine-tuned for 10 epochs using AdamW. No additional post processing is performed on the output generated by the sequence-to-sequence model.  The table reports the header F1 scores (the mean of row and column header F1 scores for the two-dimensional RotoWire datasets) and the error rate. A generated table is said to contain an error if the number of cells in any row or column of the table is inconsistent with the number on any other row or column. A table is said to be perfect if and only if all rows of the table contain an equal number of column cells and vice versa. 

\mbox{gTBLS} significantly outperforms the sequence-to-sequence baseline, with up to 3x improvement in Header F1 and 6x improvement in cell F1 for Table Construction and Table Content Generation tasks, respectively. Notably, on the RotoWire datasets, \mbox{gTBLS} excels, consistently generating valid tables while the sequence-to-sequence approach exhibits an error rate exceeding 50\% on the RotoWire Player dataset. \mbox{gTBLS} ensures the reliability of all generated tables through its two-stage process. 

\subsection{Error Propagation}
The two-stage approach of \mbox{gTBLS} raises the question of error propagation since the question-answering stage utilizes the headers generated in the first stage. 
Table \ref{tab:error_prop} presents an ablation study where the best performing question-answering model is tasked with generating cells using headers obtained from teacher-forcing (Gold headers) and predicted headers from the first stage of \mbox{gTBLS}. 

As expected, using headers from teacher-forcing outperforms using predictions. Using predicted headers achieves parity on E2E and WikiBio, with a gap $<$1\%. We posit that this is due to the relatively straightforward nature of the headers indicated by the high F1 and BERTScores in Table~\ref{tab:comapre_sota_tab_con}. The performance on WikiTableText degrades by 4\%, possibly due to variance in the dataset, with limited consistency in the presence of titles and subtitles across tables. The error propagation is highest on the two-dimensional RotoWire dataset, a combination of the fact it is relatively smaller in size (Table \ref{tab:ds_statistics}) and the two-dimensional nature, so errors across row and column headers add up.

\begin{table}[t]
\addtolength{\tabcolsep}{-1pt}
    \centering
    \begin{tabular}{c|c|c|c}
        \toprule 
        \multirow{2}{*}{Dataset} & Gold & Pred. Headers& \multirow{2}{*}{Diff (\%)} \\
        & headers & (\mbox{gTBLS}) &  \\
        \midrule
        E2E & 98.29 & 97.91 & 0.38\\
        WTT & 72.41 & 68.09 & 4.32 \\
        WikiBio & 67.45 & 67.10 & 0.35 \\
        RW - Team & 95.94 & 89.09 & 6.85 \\
        RW - Player & 88.75 & 86.09 & 2.66 \\
        \bottomrule
    \end{tabular}
    \caption{Ablation study to highlight the difference in F1 score when using headers obtained from teacher forcing versus headers predicted by the Table Content Generation network in \mbox{gTBLS}.}
    \label{tab:error_prop}
    \vspace{-10pt}
\end{table}

\section{Conclusion}
\label{sec:conclusions}
This paper introduces Generative Tables (\mbox{gTBLS}), an approach to generate tables from text. \mbox{gTBLS} uses a two-stage process, first constructing a tabular structure using a causal language modeling objective followed by question answering to fill in the content.  A key advantage of the two-stage approach is that all tables generated by \mbox{gTBLS} are valid without requiring post-processing, resulting in up to 57\% reduction in error rates when compared to sequence-to-sequence approaches.   \mbox{gTBLS} improves prior approaches by up to 20\% in BERTScore and achieves overall parity in F1 on the table content generation task on the E2E, \mbox{WikiTableText}, WikiBio, and RotoWire datasets. Furthermore, the question-answering component of \mbox{gTBLS} is modular, with billion parameter instruction fine-tuned models demonstrating performance close to fine-tuned approaches. Leveraging LLMs in a zero-shot configuration presents an approach for table generation in situations where fine-tuning is infeasible.

\section{Limitations}
\label{sec:limitations}

The \mbox{gTBLS} method, though effective for table generation from text, presents unresolved challenges. First, its performance is limited by the context length of the utilized models, leading to the omission of header and cell information from later parts of the source text. Additionally, its reliance on generating question-answer pairs from row and column headers restricts it to tables with a direct header-cell correlation. Complex table structures, like headers spanning multiple rows or columns, remain a challenge. Moreover, \mbox{gTBLS} is optimized for generating dense tables, where cell content directly corresponds to the text. This study excludes cells without matching text information to align with evaluation frameworks proposed by prior work. However, future approaches could explore generating sparse tables, potentially incorporating unknown <UNK> tokens as needed.
Finally, reducing the gap in Table \ref{tab:error_prop} is a challenge we plan on addressing in future work through the use of additional question answering to rectify erroneous headers in the first stage.



\bibliography{anthology,custom, Table_Generation}
\bibliographystyle{acl_natbib}

\appendix

\section{Appendix}
\label{sec:appendix}
We acknowledge the use of GitHub Copilot to assist in code completion.

\subsection{Dataset Examples}
\label{sec:dataset_examples}
This section details example textual paragraphs and associated tables from the different datasets. 

\flushleft{\textbf{E2E}: }
\newline The Eagle is a low rated coffee shop near Burger King and the riverside that is family friendly and is less than £20 for Japanese food.
\begin{table}[h]
    \centering
    \begin{tabular}{c|c}
        \toprule 
        Name  & The Eagle \\
        Food  & Japanese \\
        Price range & Less than £20 \\
        Customer Rating & Low \\
        Area & Riverside \\
        Family friendly & Yes \\
        Near & Burger King \\
        \bottomrule
    \end{tabular}
    \label{tab:e2e}
    \vspace{-10pt}
\end{table}

\flushleft{\textbf{WikiTableText}: }
\newline Michelle Schimel was New York State assemblywoman in Portuguese Heritage Society.
\begin{table}[h]
    \centering
    \begin{tabular}{l|l}
    \toprule 
         Title & Potuguese Heritage Society  \\
         Subtitle & Other activities \\
         Name & Michelle Schimel \\
         \bottomrule
    \end{tabular}
    \label{tab:wtt}
    \vspace{-10pt}
\end{table}

\flushleft{\textbf{WikiBio}: } 
\newline Leonard Shenoff Randle (born February 12, 1949) is a former Major League Baseball player. He was the first-round pick of the Washington Senators in the secondary phase of the June 1970 Major League Baseball draft, tenth overall.

\begin{table}[h]
    \centering
    \begin{tabular}{c|c}
       \toprule 
       Debut team  & Washington Senators \\
       Name  & Lenny Randle \\
       Birth Date & 12 February 1949  \\
       \bottomrule
    \end{tabular}
    \label{tab:Wikibio}
    \vspace{-10pt}
\end{table}

\flushleft{\textbf{RotoWire}}
\newline 
The Atlanta Hawks (46 - 12) beat the Orlando Magic (19 - 41) 95 - 88 on Friday. Al Horford had a good all - around game, putting up 17 points, 13 rebounds, four assists and two steals in a tough matchup against Nikola Vucevic. Kyle Korver was the lone Atlanta starter not to reach double figures in points. Jeff Teague bounced back from an illness, he scored 17 points to go along with seven assists and two steals. After a rough start to the month, the Hawks have won three straight and sit atop the Eastern Conference with a nine game lead on the second place Toronto Raptors. The Magic lost in devastating fashion to the Miami Heat in overtime Wednesday. They blew a seven point lead with 43 seconds remaining and they might have carried that with them into Friday's contest against the Hawks. Vucevic led the Magic with 21 points and 15 rebounds. Aaron Gordon (ankle) and Evan Fournier (hip) were unable to play due to injury. The Magic have four teams between them and the eighth and final playoff spot in the Eastern Conference. The Magic will host the Charlotte Hornets on Sunday, and the Hawks with take on the Heat in Miami on Saturday.

\begin{table}[h]
    \centering
    \begin{tabular}{c|c|c|c|c}
    \toprule 
        & \multirow{2}{*}{Losses} & \multirow{2}{*}{Total points} & Points in & \multirow{2}{*}{Wins} \\
        & & & 4th quarter & \\
        \midrule
        Magic & 41 & 88 & 21 & 19 \\
        Hawks & 12 & 95 & & 46 \\
        \bottomrule
    \end{tabular}
    \label{tab:rw_team}
    \vspace{-10pt}
\end{table}

\begin{table}[h]
\addtolength{\tabcolsep}{-1pt}
    \centering
    \begin{tabular}{c|c|c|c|c}
    \toprule 
         & Assists & Points & Rebounds & Steals  \\
         \midrule 
        Nikola Vucevic & & 21 & 15 & \\
        Al Horford & 4 & 17 & 13 & 2 \\
        Jeff Teague & 7 & 17 & & 2 \\
        \bottomrule
    \end{tabular}
    \label{tab:rw_player}
    \vspace{-5pt}
\end{table}

\newpage
\subsection{Hyperparameters}
\label{sec:hyperparams}

Header generation 
\begin{table}[h]
\addtolength{\tabcolsep}{-1pt}
    \centering
    \begin{tabular}{c|c|c|c|c|c}
    \toprule
         \multirow{2}{*}{Dataset} & \multirow{2}{*}{lr} & Batch  & \multirow{2}{*}{Warmup} & \multirow{2}{*}{Epochs} & \multirow{2}{*}{Tokens}  \\
         & & Size & & & \\
         \midrule
         WTT & 1e-4 & 32 & 1000 & 10 & 512 \\
         Wikibio & 1e-4 & 64 & 2000 & 10 & 512 \\
         E2E & 1e-4 & 128 & 250 & 10 & 256 \\
         RotoWire & 1e-4 & 32 & 250 & 10 & 512 \\
         \bottomrule
    \end{tabular}
    \caption{Hyperparameters for header generation experiments}
    \label{tab:header_gen}
    \vspace{-5pt}
\end{table}

Answer generation

\begin{table}[h!]
\addtolength{\tabcolsep}{-1pt}
    \centering
    \begin{tabular}{c|c|c|c|c|c}
             \toprule
         \multirow{2}{*}{Dataset} & \multirow{2}{*}{lr} & Batch  & \multirow{2}{*}{Warmup} & \multirow{2}{*}{Epochs} & \multirow{2}{*}{Tokens}  \\
         & & Size & & & \\
         \midrule
         WTT & 1e-4 & 128 & 300 & 10 & 256 \\
         Wikibio & 1e-4 & 256 & 5000 & 10 & 512 \\
         E2E & 1e-4 & 256 & 700 & 10 & 256 \\
         RotoWire & 1e-4 & 32 & 250 & 10 & 512 \\
         \bottomrule
    \end{tabular}
    \caption{Hyperparameters for answer generation experiments}
    \label{tab:answer_gen}
\end{table}

\end{document}